# PADSNet: Physics-Aware Dual-Stream Network with Biomechanical Priors for Laryngeal Lesion Classification

Zhenzhen You, Weifeng Yu, Zhenghao Shi, Shuangli Du, Minghua Zhao, Yingcheng Zhao, Xinhong Hei, Demin Kong, Xiaoyong Ren, Jin Hou

***Abstract*—Precise classification of laryngeal lesions is clinically crucial yet highly challenging, as inter-class visual differences are often subtle. While abnormalities in vocal fold vibratory dynamics provide complementary discriminative information, existing deep learning models predominantly rely on visual appearance and lack explicit mechanisms for modeling clinically meaningful vibratory dynamics. To address this, we propose PADSNet, a Physics-Aware Dual-Stream Network for laryngeal lesion classification that integrates visual representations with explicit biomechanical reasoning. Specifically, a Wavelet-Attention Enhancement Module (WAEM) is designed to bridge the spectral-semantic gap, enabling complementary learning of spatial texture and dynamic frequency representations. Meanwhile, a Knowledge Computing Module (KCM) extracts interpretable biomechanical indices from temporal glottal distance waveforms, including indicators of mobility, symmetry, glottal closure, and vibration stability. These descriptors are subsequently incorporated into a Physics-Aware Gating Module (PAGM) as biomechanical priors to adaptively recalibrate visual feature toward pathology-relevant vibratory patterns. Extensive experiments on a clinical dataset demonstrate that PADSNet achieves an accuracy of 0.8452, an F1-score of 0.8090, a sensitivity of 0.7953, and a specificity of 0.8866. Notably, the F1-score outperforms state-of-the-art video models by a margin of 2.0% to 4.0%. These competitive baselines include convolution-based video models (e.g., X3D), Transformer-based video models (e.g., Video Swin Transformer), self-supervised video representation models (e.g., VideoMAE and V-JEPA 2), state-space video models (e.g., VideoMambaPro), and large-scale or domain-specific video foundation models (e.g., InternVideo2 and SurgMotion). These results indicate that explicitly integrating biomechanical priors with deep visual representations provides a more accurate, robust, and clinically interpretable framework for automated laryngeal lesion classification.**

## I. INTRODUCTION

ACCORDING to GLOBOCAN 2022 estimates reported in 2024 Global Cancer Statistics, laryngeal cancer remains a substantial global health burden with approximately 189,000 new cases worldwide in 2022[1]. The prognosis for patients depends strongly on the stage at diagnosis, which in turn relies on accurate assessment of laryngeal lesions during the initial presentation.

In outpatient settings, laryngoscopy is the primary tool for evaluating laryngeal lesions providing direct visualization of vocal fold morphology. However, accurate interpretation of laryngoscopic findings remains challenging because the inter-class visual differences among diagnostic categories are often subtle and uneven. As shown in Fig. 1a, high-risk, low-risk and normal tissue cases may exhibit partially overlapping static appearances in terms of color, texture, surface morphology, or glottal configuration. This makes assessment subjective and prone to inter-observer variability[2][3].

Stroboscopy can alleviate this limitation by slowing high-speed vocal fold motion to reveal vibratory patterns of the mucosal wave, which provide complementary cues for discrimination. The corresponding glottal distance waveform in Fig. 1b further illustrates that cases with visually similar static appearances can still differ in these dynamic features such as mucosal wave asymmetry, reduced periodicity, and irregular glottal opening-closing patterns. These observations indicate that such dynamic cues are not merely supplementary but can be decisive when static appearance is ambiguous, and should therefore be incorporated into automated lesion differentiation. Nevertheless, quantitative frame-by-frame analysis of vocal fold vibration is labor-intensive and impractical in routine outpatient examinations [4][5], creating a pressing need for an automated pipeline capable of objectively characterizing glottal dynamics and assisting clinical decision-making.

This work was supported in part by the National Natural Science Foundation of China (NSFC) [62502377, 62571425]; Key Research and Development Program of Shaanxi Province [2026SF-YBXM-471]; Scientific Research Project of Shaanxi Provincial Department of Education [25JR127]; Xi'an Science and Technology Plan Project [25GXKJRC00010]; Natural Science Basic Research Program of Shaanxi [2025JC-YBMS-755]; China University Industry-Academia-Rsearch Innovation Fund-Next Generation Information Technology Innovation Project [2024IT088]; the Open Research Program of La-boratory for Microwave Spatial Intelligence and Cloud Platform [2025ZY01019]; Doctoral Dissertation Innovation Fund of Xi'an University of Technology [BC202621]. (*Corresponding authors: Jin Hou*).

Zhenzhen You, Weifeng Yu, Zhenghao Shi, Shuangli Du, Minghua Zhao, Yingcheng Zhao, and Xinhong Hei are with the Human-Machine Integration Intelligent Robot Shaanxi University Engineering Research Center, School of Computer Science and Engineering, Xi'an University of Technology, Xi'an, China (e-mail: zhenzhen_you@xaut.edu.cn, 2241221076@stu.xaut.edu.cn, ylshi@xaut.edu.cn, dusl@xaut.edu.cn, zhaominghua@xaut.edu.cn, zyc@xaut.edu.cn, heixinhong@xaut.edu.cn).

Demin Kong, Xiaoyong Ren, and Jin Hou are with Department of Otolaryngology, Second Affiliated Hospital of Medical College, Xi'an Jiaotong University, Xi'an, China (e-mail: 1803047714@qq.com, renxiaoyong@mail.xjtu.edu.cn, dochou@xjtu.edu.cn).

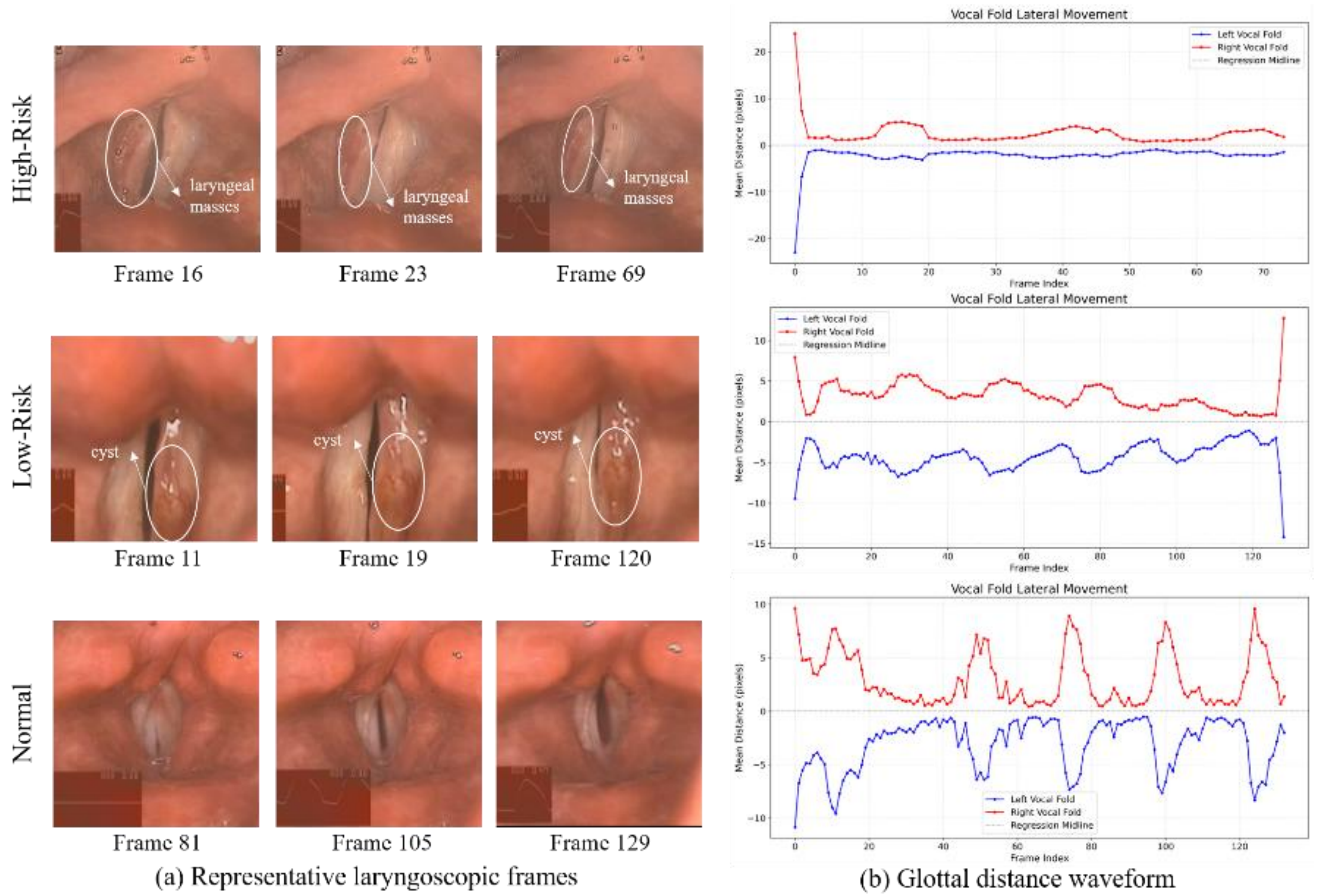

Fig. 1 (a) Representative stroboscopic laryngoscopic frames, illustrating subtle inter-class visual differences among high-risk, low-risk and normal tissue categories. (b) Glottal distance waveforms derived from the computed glottal midline (defined as the symmetry axis of the glottal area). The two curves represent the distance variations of the left and right vocal fold relative to this midline over time.

Despite this need, automated classification of laryngeal lesions remains highly challenging. First, unlike general video recognition tasks, where different classes are usually characterized by distinct appearance or motion patterns, laryngoscopic lesion classification involves much finer diagnostic boundaries [6][7]. For many borderline cases, morphological differences are minimal, while the decisive evidence is more closely related to biomechanical abnormalities such as increased stiffness, impaired mucosal wave propagation, or irregular vibration. Consequently, models relying primarily on visual appearance are prone to misclassfication when distinguishing clinically similar lesions. Second, this intrinsic inter-class ambiguity is further exacerbated by the acquisition characteristics of laryngoscopic videos. As shown in Fig. 2, specular reflections from moist mucosal surfaces, illumination fluctuations, physiological occlusions, and motion blur caused by rapid vocal fold vibration can obscure fine mucosal textures and boundary details [6][7][8]. More importantly, these artifacts tend to mask precisely the subtle vibration-sensitive cues that are most critical for identifying early abnormalities. Therefore, the central challenge is not merely to recognize visually abnormal lesions, but to detect weak pathological deviations from normal vocal fold vibration under severe visual interference and limited inter-class separability.

In recent years, deep learning has substantially advanced automated laryngoscopic image analysis, progressing from frame-based CNNs [9] and recurrent architectures [10] to modern video Transformers such as the Video Swin Transformer (VST) [11] . These methods provide increasingly powerful visual and spatiotemporal representations; however, their application to laryngeal lesion classification remains predominantly data-driven. Although video models can implicitly encode vocal-fold motion, they do not explicitly incorporate clinically interpretable biomechanical information such as bilateral symmetry, glottal closure, periodicity, and vibration stability. This limitation may be particularly relevant when diagnostic categories exhibit similar visual appearances and subtle differences are reflected in vocal-fold dynamics. These observations motivate a physics-aware framework that complements learned spatiotemporal representations with explicit vocal-fold biomechanical priors [12].

To address the accurate classification of laryngeal lesions and bridge the gap between video representations and clinical biomechanical reasoning, this paper proposes a novel Physics-Aware Dual-Stream Network, termed PADSNet. Our core hypothesis is that laryngeal lesion classification should be treated as a biomechanical discrimination problem in addition to a conventional visual recognition task. By incorporating

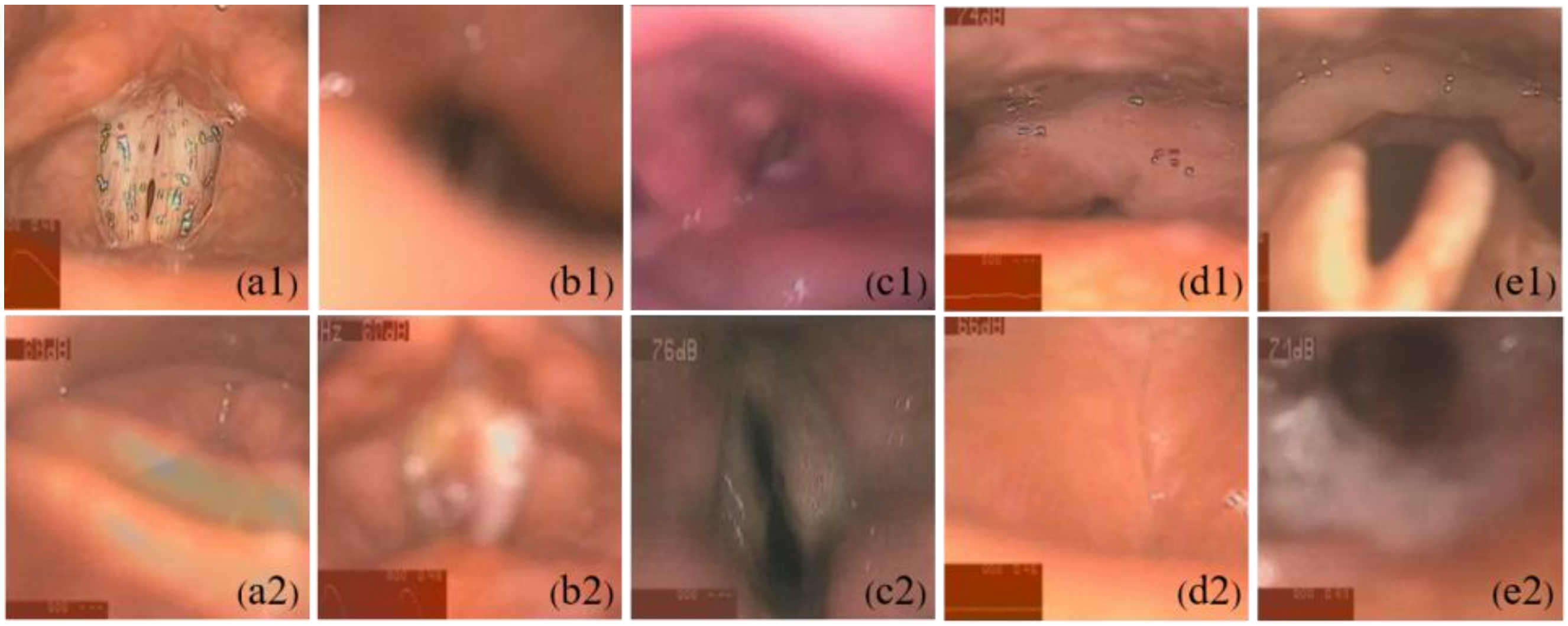

Fig. 2 (a1)-(a2) specular reflections; (b1)-(b2) motion blur; (c1)-(c2) illumination and varying color temperatures; (d1)-(e2) physiological occlusions.

explicit vocal fold dynamic priors, the model can be guided to enlarge the effective discriminative margin between clinically similar cases, where visual appearance differences are minimal but subtle vibratory abnormalities may already exist.

Specifically, PADSNet contains two complementary streams. The visual stream adopts the VST to extract hierarchical spatiotemporal appearance representations from laryngoscopic video clips, while the biomechanical stream models the 2-dimensional representation of glottal distance extracted from vocal fold segmentation masks. This representation provides a compact and physically meaningful description of vocal fold vibration, allowing the network to capture early deviations in symmetry, periodicity, closure, and stability. To facilitate effective interaction between these heterogeneous representations, we introduce WAEM. Instead of performing direct feature fusion, this module decomposes visual and biomechanical features into low-frequency trend components and high-frequency detail components, enabling band-wise cross-modal interaction between slowly varying trends and rapidly varying temporal components.

Furthermore, we design a PAGM to explicitly inject interpretable biomechanical indices extracted by the KCM into the final decision process. These indices serve as clinical inductive biases that dynamically recalibrate deep features toward pathology-relevant vibration patterns. Specifically, periodicity and waveform consistency characterize regular vocal fold vibration, while asymmetry, incomplete glottal closure, reduced mobility, increased gap-related changes, and irregular vibration provide complementary cues for distinguishing different lesion categories. Through this mechanism, PADSNet is encouraged to focus not only on visually salient regions but also on physically meaningful dynamic deviations, thereby improving lesion differentiation under weak inter-class visual separability.

The main contributions of this paper are summarized as follows:

- We emphasize laryngeal lesion classification as a biomechanically informed diagnostic problem, with particular attention to the clinically challenging boundary between clinically similar cases.
- We propose PADSNet, a physics-aware dual-stream framework that integrates spatiotemporal appearance features with glottal waveform-derived biomechanical representations for detecting early pathological deviations.
- We introduce a WAEM to facilitate band-wise cross-modal interaction between low-frequency trend and high-frequency detail representations of the visual and biomechanical streams.
- We develop a PAGM that uses interpretable biomechanical indices to recalibrate deep features around clinically meaningful physical patterns, such as periodicity, symmetry, glottal closure, and vibration stability.
- Extensive experiments on a clinical laryngoscopy video dataset demonstrate that PADSNet outperforms representative video understanding baselines, validating the effectiveness of biomechanical priors under weak inter-class visual separability.

## II. Related works

### A. Vocal Cord Lesion Classification

Early studies on vocal cord lesion classification mainly relied on handcrafted descriptors extracted from static laryngoscopic images[13][14][15]. These methods encoded color, texture, and geometric patterns using features such as Histograms of Oriented Gradients (HOG)[16], Local Binary Patterns (LBP)[17], and Scale-Invariant Feature Transform (SIFT) [18], followed by conventional classifiers such as Support Vector Machines (SVM) [19] or k-Nearest Neighbors (KNN)[20]. Although these approaches provided preliminary evidence for computer-aided vocal fold lesion recognition, their performance was constrained by labor-intensive feature engineering and the limited representational capacity of shallow descriptors.

Beyond static appearance analysis, quantitative vocal-fold dynamics were also investigated in earlier work. Voigt et al. [21]

derived phonovibrogram-based descriptors from high-speed laryngeal recordings to characterize bilateral vibratory behavior and used an SVM to classify healthy and dysfunctional vibration patterns. This study demonstrated the diagnostic value of vocal-fold dynamics, but such approaches largely relied on manually designed descriptors and conventional classifiers.

Recent deep-learning studies have achieved strong performance in laryngoscopic image and video classification under different diagnostic settings. Cho et al.[22] analyzed 4,106 images covering eight common laryngeal disorders and normal cases, reporting micro- and macro-average PR-AUCs of 0.95 and 0.91, respectively. Tran et al. [23] evaluated 4,549 flexible-laryngoscopy images from 876 patients for three-class vocal-fold status classification and reported a best overall accuracy of 97.51% using Xception. Wei et al. [24] evaluated a ResNet–Transformer dual-stream model on the 3,057-image Laryngoscope8 dataset covering eight laryngeal categories and achieved 86.51% accuracy with a mean AUC of 0.954. At the video level, Qiu et al. [25] analyzed 1,109 laryngoscopic videos for laryngeal cancer detection and reported 92.4% accuracy and 94.8% F1.

Despite these advances, most deep-learning methods remain driven by visual or spatiotemporal representations, whereas earlier vibration analysis offers interpretable but shallow descriptors. These two lines have evolved largely in parallel, and bridging them is non-trivial because vocal-fold dynamics are subtle, high-frequency, and tightly coupled with appearance. This gap remains open for reliable laryngeal lesion classification.

### *B. Spatiotemporal Modeling Paradigms*

Spatiotemporal modeling is essential for video-based medical analysis because diagnostic information is often embedded in both static appearance and temporal dynamics. In general video understanding, representative paradigms include convolution-based video models, Transformer-based models, self-supervised video representation learning, large-scale video foundation models, and state-space models. These paradigms provide useful references for evaluating whether general-purpose video backbones can be directly applied to laryngoscopic lesion classification.

CNN-based video models, such as X3D[26], extend image-level convolutional networks to the temporal domain by learning local spatiotemporal patterns from consecutive frames[27]. Owing to their efficient convolutional design, these models can capture short-range motion cues with relatively low computational cost. Transformer-based video models further improve temporal modeling through self-attention mechanisms [28] enabling the network to capture longer-range dependencies and more flexible interactions across space and time. For example, VST [11] introduces hierarchical shifted-window attention into the spatiotemporal domain, allowing the model to jointly encode local motion patterns and global temporal context. In endoscopic image analysis, Kusters et al. systematically compared CNN- and Transformer-based architectures in terms of performance, robustness, and generalization[29], while Wang et al. proposed TSdetector to improve colonoscopy video detection through temporal consistency and spatial reliability learning[30]. However, these studies mainly focus on appearance-driven endoscopic analysis or lesion detection and have not addressed larynx-specific vibratory biomechanics.

More recently, self-supervised and foundation video models have shown strong transferability by learning generic video representations from large-scale data. VideoMAE [31] adopts masked video reconstruction to learn robust spatiotemporal features, while V-JEPA2[32] follows a joint-embedding predictive learning paradigm to capture higher-level temporal dynamics. InternVideo2 [33] further leverages large-scale video pretraining and multimodal supervision to obtain general-purpose video representations, and SurgMotion [34] extends foundation-model learning to surgical videos by emphasizing motion-oriented representation learning. In addition, structured state-space models such as Video Mamba Pro [35] provide an efficient alternative for long-sequence video modeling, making them attractive for long-duration medical videos.

General-purpose video models are optimized for salient appearance and coarse motion, whereas diagnostic cues in stroboscopy are weak, periodic, and often obscured by acquisition artifacts. Simply scaling up a stronger backbone therefore does not address the core difficulty. How to bias spatiotemporal representations toward vocal-fold biomechanics remains an open question.

## III. Methods

### *A. Overview*

As illustrated in Fig. 3, PADSNet employs a dual-branch architecture specifically designed to process heterogeneous representations derived from a single laryngoscopic video source. Its core motivation lies in bridging the gap between two distinct types of signals: high-dimensional, implicitly learned visual cues and low-dimensional, physically meaningful dynamic signals. Given an input video sequence, the framework diverges into two parallel streams. (1) The visual stream utilizes the VST to extract hierarchical spatiotemporal features, capturing complex inter-frame visual evolution and texture dynamics. (2) The biomechanical stream processes the 2-dimensional representation of glottal distance extracted from the same video. It employs a Bidirectional LSTM[36] to capture implicit dynamic embeddings while leverages a KCM to obtain explicit biomechanical indices. These streams are subsequently integrated via two key modules: a WAEM and a PAGM. The WAEM addresses feature-level spectral-semantic misalignments across different frequency bands, while the PAGM explicitly utilizes biomechanical prior information to recalibrate deep features towards clinically relevant vibration anomalies. Finally, the fused features are fed into a classification head for disease prediction.

### *B. Preprocessing*

To reduce the influence of non-diagnostic intervals and visual artifacts, we adopted a coarse-to-fine preprocessing pipeline. Each video was first decoded into frame sequences, from which anchor frames were sampled at intervals of 50 frames. Frames with severe overexposure, under-exposure, motion blur, or non-laryngeal views were removed based on brightness statistics and Laplacian variance[37]. The remaining frames were segmented using a pretrained nn-UNet[38], and valid phonation intervals were identified according to the glottal mask area and anatomical plausibility.

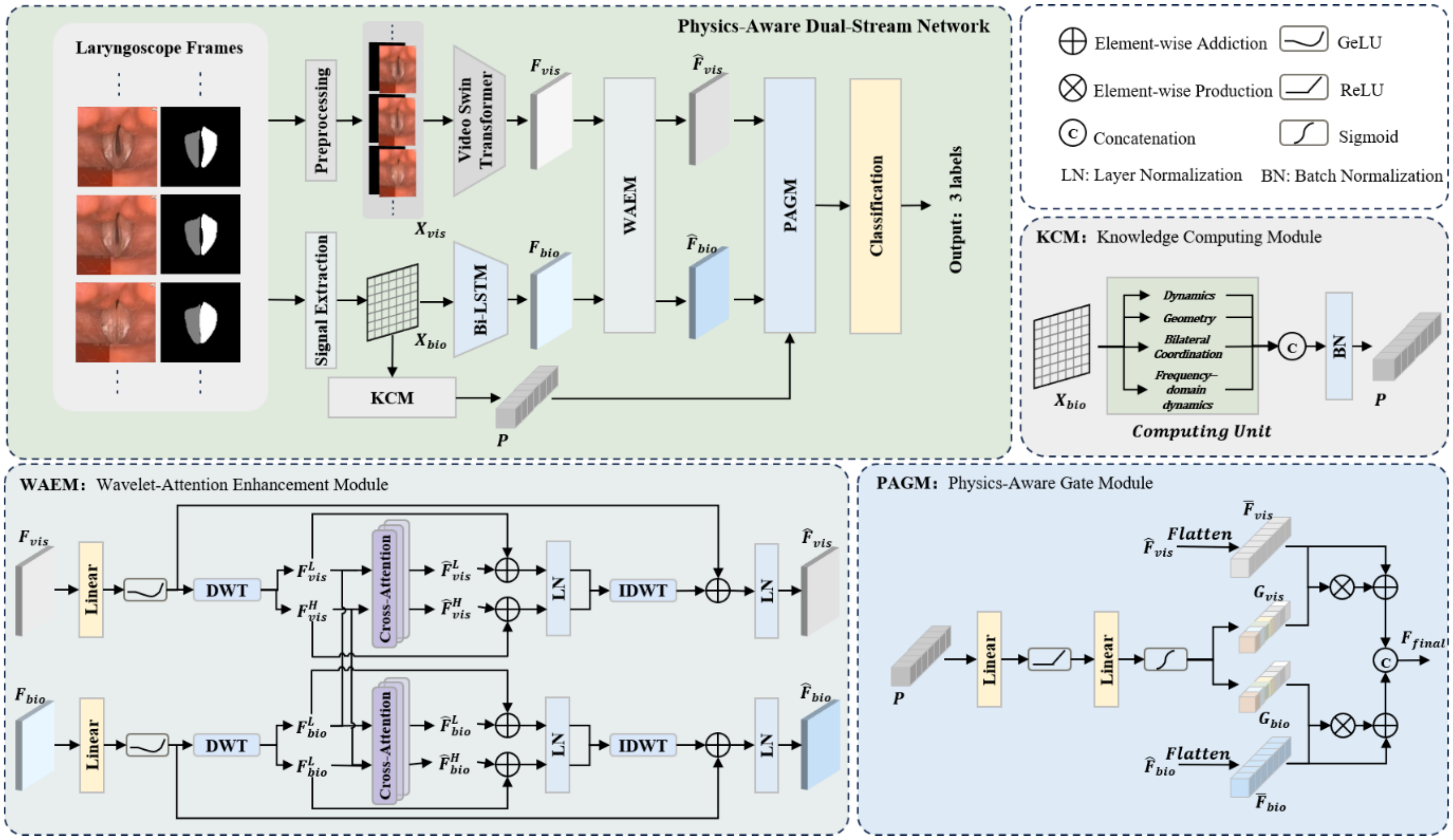

Fig. 3. PADSNet Architecture Overview. The visual stream extracts spatiotemporal features using a Video Swin Transformer (VST), while the biomechanical stream models bilateral glottal-distance sequences with a Bi-LSTM. The KCM derives explicit biomechanical priors from vocal-fold dynamics. WAEM performs frequency-aware interaction between the two learned feature streams, and PAGM uses the biomechanical priors to adaptively recalibrate the fused representations before classification.

For the training set, the localized phonation interval was further refined to identify a stable vibration segment. Frame-wise left and right glottal distances were extracted from the vocal-fold masks. Frames with invalid masks or distance estimates were excluded, while excessively open states were detected using an adaptive threshold derived from the distribution of glottal distances. Short interruptions were suppressed by a three-frame binary closing operation, and contiguous valid segments shorter than 15 frames were discarded.

The remaining candidate segments were ranked using a label-independent quality score combining normalized segment duration, Laplacian-based image sharpness, vibration periodicity estimated from the autocorrelation of the glottal-width sequence, and bilateral symmetry measured from the relative difference between the mean left and right glottal distances. These four components were weighted by 0.35, 0.30, 0.25, and 0.10, respectively. A lower weight was assigned to the symmetry term to retain potentially informative asymmetric vibration patterns. The highest-scoring candidate was retained for training.

During testing, the above micro-segment selection strategy was not applied. Instead, each video was densely sampled using a sliding window over the valid phonation interval, and clip-level softmax probabilities were averaged to obtain the final video-level prediction. This inference strategy avoids artificially selecting an optimal test segment and better reflects realistic clinical deployment.

For each sampled clip, RGB frames were cropped around the glottal region, resized, and normalized. The corresponding binary masks were concatenated with RGB frames along the channel dimension, forming a four-channel input tensor $\boldsymbol{X}_{vis} \in \mathbb{R}^{T\times H\times W\times 4}$, where $T = 32$ denotes the number of frames in the clip, $H = W = 224$ denote the height and width of each resized frame, and the four channels correspond to the RGB channels and the binary glottal mask.

### *C. Biomechanical Signal Extraction*

To obtain a physiological meaningful reference for quantifying vocal fold lateral excursion while suppressing distortions caused by irregular lesion morphology, we extract a 2-dimensional representation of glottal distance directly from the segmentation masks and propose an anchor-based robust regression strategy to determine the stable glottal midline (Fig.4). Specifically, for each frame $t$, we scan the segmentation mask row by row. Let $x_L$ and $x_R$ denote the $x$-coordinates of the inner edges of the left and right vocal cords at a given row, respectively. The instantaneous midline point at that row is calculated as $x_m = (x_L + x_R)/2$.

The anterior and posterior commissures provide relatively stable anatomical references for glottal midline estimation. Based on clinical consultation, laryngeal lesions most commonly involve the mid-membranous portion of the vocal folds, where local morphological deformation may distort midpoint estimation. Accordingly, we define an anchor set $\mathbb{S}_{anchor}$, comprising only the detected midpoints from the anterior $\alpha$% and posterior $\alpha$% regions (clinically set to $\alpha$=25), effectively filtering out the potentially distorted central region.

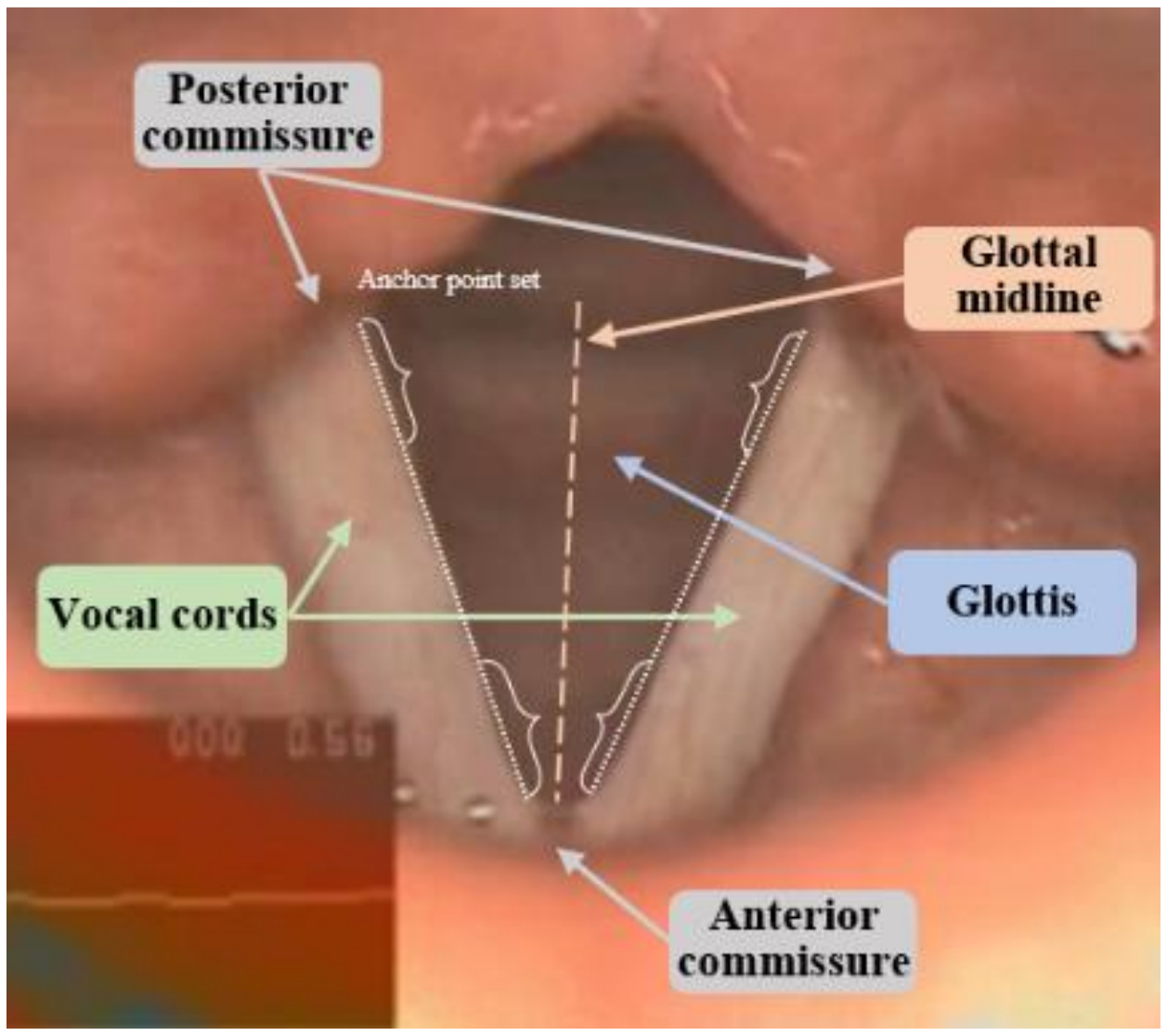

Fig. 4 Anatomical landmarks of the glottal region under clinical laryngoscopy. The anterior and posterior commissures serve as stable physiological anchor points, while the left and right vocal folds constitute the primary vibrating boundaries.

To further suppress segmentation outliers, we fit the glottal midline $\mathcal{L}$ to the points in $\mathbb{S}_{anchor}$ by minimizing the Huber loss[39]. Assuming the midline can be modelled as a linear equation $\boldsymbol{x} = k\boldsymbol{y} + b$, the optimal parameters are found by

$$\min_{k,b} \sum_{(x_i,y_i)\in\mathbb{S}_{anchor}} H_\delta(x_i - (ky_i + b)) \quad (1)$$

where $H_\delta(\cdot)$ is the Huber loss function, which exhibits a quadratic relationship for small errors and a linear relationship for large errors, ensuring robustness against segmentation noise.

Finally, we calculate the average lateral Euclidean distance from all edge points of the left and right vocal cords to this robust midline $\mathcal{L}$, producing two displacement sequences:

$$\boldsymbol{d}_L = [d_L(1), d_L(2), \ldots, d_L(\mathrm{T})] \quad (2)$$

$$\boldsymbol{d}_R = [d_R(1), d_R(2), \ldots, d_R(\mathrm{T})], \quad (3)$$

where $d_L(\mathrm{T})]$ and $d_R(\mathrm{T})$ denote the lateral displacements of the left and right vocal folds at frame $t$, respectively. The corrected bivariate biomechanical time series is then formulated as:

$$\boldsymbol{X}_{bio} = [\boldsymbol{d}_L, \boldsymbol{d}_R] \in \mathbb{R}^{T\times 2}. \quad (4)$$

This representation characterizes vocal cords lateral excursion relative to the estimated physiological axis while reducing sensitivity to lesion-induced morphological asymmetries.

Building upon the biomechanical time-series signals, we employed KCM to derive a set of handcrafted biomechanical indicators. The KCM was designed to provide a compact yet complementary description of vocal-fold vibration rather than to optimize an arbitrary latent dimensionality. Specifically, the selected descriptors cover four clinically relevant aspects of vocal-fold behavior: overall mobility, glottal geometry, bilateral coordination, and frequency-domain regularity. These aspects are represented by one mobility descriptor, three gap-related descriptors, two bilateral coordination descriptors, and three frequency-domain descriptors, respectively, resulting in a 9-dimensional prior vector $\boldsymbol{P} \in \mathbb{R}^9$. The dimension is therefore determined by the selected set of complementary biomechanical descriptors rather than by an independently tuned embedding size. Additional descriptors could be incorporated, but were not included to avoid introducing redundant measures that capture information already represented by the existing indices.

**1) Dynamics**

To differentiate between immobile (e.g., paralyzed) and vibrating vocal folds, we calculate the mobility index. Let $\boldsymbol{v}_L$ and $\boldsymbol{v}_R$ denote the first-order temporal differences (i.e., velocities) of the left and right distance sequences, respectively. This index is defined as the arithmetic mean of their standard deviations:

$$D = \frac{1}{2}(\sigma(\boldsymbol{v}_L) + \sigma(\boldsymbol{v}_R)), \quad (5)$$

where low value of mobility index clearly indicates the presence of paralyzed or fixed lesions.

**2) Geometry**

To characterize glottal closure competence, we analyze the instantaneous glottal gap sequence

$$g(t) = d_L(t) + d_R(t). \quad (6)$$

Based on this sequence, three complementary descriptors are extracted.

*Mean Gap:* The mean glottal gap reflects the average separation width between the bilateral vocal folds during phonation:

$$\mu_g = \frac{1}{T}\sum_{t=1}^{T} g(t). \quad (7)$$

*Minimum Gap:* The minimum glottal gap serves as a proxy for glottal closure competence:

$$g_{min} = min_{1\le t\le T}\, g(t). \quad (8)$$

A larger non-zero value may indicate incomplete glottal closure.

*Gap Standard Deviation:* The temporal fluctuation of the glottal gap is measured by the standard deviation of $g(t)$, capturing the temporal variability of glottal opening and closing during phonation.

**3) Bilateral Coordination**

Lesions often disrupt the coordinated vibration of the bilateral vocal folds. Therefore, bilateral coordination is quantified from both symmetry and waveform similarity.

*Asymmetry:* To identify unilateral lesions that disrupt vibratory synchronization, we measure amplitude balance and shape consistency, calculating the asymmetry index to quantify the amplitude balance

$$A = \frac{|\sigma_L - \sigma_R|}{\sigma_L + \sigma_R + \epsilon}, \quad (9)$$

where $\sigma_L$ and $\sigma_R$ denote the standard deviations of the left and right distance sequences respectively, and $\epsilon$ is a small constant for numerical stability. This index is scale-invariant and highly sensitive to mass effects.

*Correlation:* Waveform similarity is evaluated using the cosine similarity between the mean-centered left and right waveforms

$$C_{corr} = \frac{\widetilde{\boldsymbol{d}}_L \cdot \widetilde{\boldsymbol{d}}_R}{\|\widetilde{\boldsymbol{d}}_L\|\|\widetilde{\boldsymbol{d}}_R\|}, \tag{10}$$

where $\mu_L$ and $\mu_R$ represent their respective temporal means. Subtracting the mean yields the zero-mean sequences $\widetilde{\boldsymbol{d}}_L = \boldsymbol{d}_L - \mu_L$ and $\widetilde{\boldsymbol{d}}_R = \boldsymbol{d}_R - \mu_R$ .This mean-centring operation is crucial as it isolates the dynamic vibratory component by removing the static anatomical offset. Consequently, it ensures that the cosine similarity strictly evaluates the morphological and phase synchronization of the mucosal waves, entirely independent of the absolute glottal gap.

#### 4) Frequency-domain dynamics

Frequency-domain characteristics were computed from the bilateral glottal-distance sequences using the Fast Fourier Transform (FFT). For each input clip, FFT was applied to the left and right distance sequences over a fixed temporal window of $T = 32$ consecutive frames. The DC component was discarded, and the non-DC spectral components were subsequently used to derive spectral magnitude and periodicity descriptors.

*Energy:* The spectral energy measures the average non-DC frequency magnitude of the left and right vocal fold distance sequences:

$$E = \frac{1}{2}\left(\sum_{k=1}^{K}(|\mathcal{F}(\boldsymbol{d}_L)_k| + |\mathcal{F}(\boldsymbol{d}_R)_k|)\right), \tag{11}$$

Here, $\mathcal{F}$ represents the FFT operation, and $K$ is the total number of frequency components. The summation starts from $k = 1$ to exclude the DC component, which mainly describes the static anatomical offset rather than dynamic vibration.

*Periodicity:* The periodicity index is defined as the average ratio between the dominant non-DC frequency magnitude and the total non-DC spectral magnitude of the two sides:

$$P_{per} = \frac{1}{2}\left(\frac{max_{k\geq1}|\mathcal{F}(\boldsymbol{d}_L)_k|}{\sum_{k=1}^{K}|\mathcal{F}(\boldsymbol{d}_L)_k| + \epsilon} + \frac{max_{k\geq1}|\mathcal{F}(\boldsymbol{d}_R)_k|}{\sum_{k=1}^{K}|\mathcal{F}(\boldsymbol{d}_R)_k| + \epsilon}\right), \tag{12}$$

with $\epsilon$ used for numerical stability.

*ZCR:* The Zero-Crossing Rate (ZCR)[40] is computed on the mean-centered signal to quantify the roughness and stochastic complexity of the vibrations.

The resulting 9-dimensional prior vector is normalized using Batch Normalization to account for differences in scale among the descriptors before being incorporated into the subsequent gating module.

### D. Dual-Stream Spatiotemporal Backbones

To capture long-range spatiotemporal dependencies and subtle texture evolution, we employ the VST as the visual encoder. Leveraging its hierarchical architecture and 3D shifted window attention mechanism, VST is inherently well-suited for this task, effectively balancing local feature extraction with global context modeling. Formally, VST processes $\boldsymbol{X}_{vis}$ and outputs a dense spatiotemporal feature map $\boldsymbol{F}_{vis} \in \mathbb{R}^{T'\times H'\times W'\times C_v}$ , where $T' = 16$ , $H' = W' = 7$ denote the downsampled spatiotemporal dimensions, and $C_v = 768$ is the channel depth of the visual stream.

Although $\boldsymbol{X}_{bio}$ is physically synchronized with the video, it constitutes a heterogeneous representation (low-dimensional dynamics for $\boldsymbol{X}_{bio}$ versus high-dimensional appearance for $\boldsymbol{X}_{vis}$). To model the temporal dependencies and extract implicit dynamic embeddings from $\boldsymbol{X}_{bio}$ , we utilize a Bidirectional Long Short-Term Memory (Bi-LSTM) network. The output is a sequence of dynamic embeddings $\boldsymbol{F}_{bio} \in \mathbb{R}^{T\times C_b}$ , where $T$ denotes the number of frames in the clip, and $C_b = 128$ is the channel depth of the biomechanical stream. To synchronize the temporal resolution with the visual features, we apply one-dimensional temporal convolution to align the temporal dimension of $F_{bio}$ from $T$ to $T'$ , matching the temporal granularity of $\boldsymbol{F}_{vis}$.

### E. Wavelet-Attention Enhancement Module

Direct fusion of $\boldsymbol{F}_{vis}$ and $\boldsymbol{F}_{bio}$ may mix heterogeneous frequency components. WAEM therefore applies Discrete Wavelet Transform (DWT)[41] to decompose both streams into low-frequency trend and high-frequency detail components, enabling band-wise cross-modal interaction before reconstruction with Inverse Discrete Wavelet Transform (IDWT). Formally, a one-level Haar DWT is applied along the temporal dimension to both feature streams, producing one low-frequency approximation subband and one high-frequency detail subband using the standard Haar coefficients $\frac{1}{\sqrt{2}}$ and $-\frac{1}{\sqrt{2}}$, these subbands represent relative temporal components. This decomposes each stream into low-frequency (trend, $L$) and high-frequency (detail, $H$) components. For each modality $m \in \{vis, bio\}$:

$$\boldsymbol{F}_m^L, \boldsymbol{F}_m^H = DWT(\boldsymbol{F}_m). \tag{13}$$

Here, $\boldsymbol{F}_{vis}^L$ captures slowly varying or persistent visual patterns, whereas $\boldsymbol{F}_{vis}^H$ emphasizes rapid frame-to-frame visual changes. Similarly, $\boldsymbol{F}_{bio}^L$ describes the overall trend of glottal opening and closing, while $\boldsymbol{F}_{bio}^H$ reflects rapid waveform fluctuations and irregular vibration.

To achieve precise semantic alignment, we employ a Multi-Head Cross-Attention (MHCA)[42] mechanism in parallel within corresponding frequency bands. Cross-attention is performed within corresponding frequency bands, allowing the low-frequency and high-frequency components of the visual and biomechanical streams to interact separately. We formulate this bidirectional interaction generally. For each frequency band $b \in \{L, H\}$, the enhanced feature $\widehat{F}_m^b$ is computed as

$$\widehat{\boldsymbol{F}}_m^b = MHCA(Q = \boldsymbol{F}_m^b, K = \boldsymbol{F}_m^b, V = \boldsymbol{F}_m^b) + \boldsymbol{F}_m^b. \tag{14}$$

This unified equation represents the four distinct interactions, effectively facilitating mutual information exchange between these heterogeneous representations.

The enhanced spectral components are recombined using the IDWT to recover the time-domain representations. For $m \in \{vis, bio\}$:

$$\widehat{\boldsymbol{F}}_m = IDWT(\widehat{\boldsymbol{F}}_m^L, \widehat{\boldsymbol{F}}_m^H) + \boldsymbol{F}_m, \tag{15}$$

we add a global residual connection to stabilize training.

### F. Physics-Aware Gating Module

While the fused feature $\widehat{\boldsymbol{F}}_m$ contains rich spatiotemporal information, it may still lack explicit physical grounding. To bridge this gap, we introduce a PAGM that dynamically modulates deep features using the explicit biomechanical prior $\boldsymbol{P}$. Specifically, the $\boldsymbol{P}$ is first projected by a two-layer Multi-Layer Perceptron (MLP), which expands the feature dimension from 9 to 64 and subsequently to 512, followed by a sigmoid activation. The resulting 512-dimensional gating vector is then divided into two 256-dimensional channel-wise attention gate $\boldsymbol{G}_{vis}$ and $\boldsymbol{G}_{bio}$:

$$[\boldsymbol{G}_{vis}, \boldsymbol{G}_{bio}] = sigmoid(MLP(\boldsymbol{P})). \quad (16)$$

After obtaining the modality-specific gates $\boldsymbol{G}_{vis}$ and $\boldsymbol{G}_{bio}$, the enhanced $\widehat{\boldsymbol{F}}_{vis}$ and $\widehat{\boldsymbol{F}}_{bio}$ are first flattened before feature recalibration. The two flattened features are then recalibrated using their corresponding gates in a residual manner:

$$\boldsymbol{F}_{final} = Concat(\overline{\boldsymbol{F}}_{vis} \odot \boldsymbol{G}_{vis} + \overline{\boldsymbol{F}}_{vis}, \overline{\boldsymbol{F}}_{bio} \odot \boldsymbol{G}_{bio} + \overline{\boldsymbol{F}}_{bio}), \quad (17)$$

where $\odot$ denotes channel-wise multiplication, $\overline{\boldsymbol{F}}_{vis}$ and $\overline{\boldsymbol{F}}_{bio}$ denote the flattened visual and dynamic feature representations, respectively. This mechanism ensures that the final classification is explicitly conditioned on clinically interpretable biomechanical indicators, effectively constraining the network's search space within a physically meaningful manifold.

The final physics-aware feature vector $\boldsymbol{F}_{final}$ is input into the Multi-Layer Perceptron (MLP) classification head to obtain the predicted probability distribution.

### *G. Loss Function*

To address the inherent class imbalance in clinical laryngoscopy datasets, we employ the Adaptive Weighted Balancing (AWB) loss [43] to train our network end-to-end. AWB loss dynamically balances training by jointly considering class frequency and prediction confidence. It comprises two components: a frequency-weighted cross-entropy term $\mathcal{L}_{ce}$ and a confidence dispersion regularization term $\mathcal{L}_{disp}$.Let $\mathcal{B}$ denote a mini-batch of size $\mathcal{B}$. For each class $k \in \{1, \ldots, C\}$, let $N_k$ be the number of samples in $\mathcal{B}$ belonging to class $k$, and let $N_{max} = max(N_k)$ . The adaptive balancing weight $\omega_k$ is defined as

$$\omega_k = \log(\frac{N_{max}}{N_k} + 1). \quad (18)$$

This logarithmic weight assigns higher importance to minority classes within the current batch.

#### 1) Adaptive Weighted Cross-Entropy Term

This term calculates the weighted average of the log probabilities for the target class. Let $p_{i,k}$ represent the predicted probability that sample $i$ belongs to its ground-truth class $k$. The mean log-probability for class $k$ is $\overline{q}_k$ as

$$\overline{q}_k = \frac{1}{N_k} \sum_{i \in \mathcal{B}_k} log(p_{i,k} + \epsilon), \quad (19)$$

where $\mathcal{B}_k$ is set of indices of samples belonging to class k in the mini-batch, and $\epsilon$ is a small scalar used to ensure numerical stability by preventing undefined logarithms. The $\mathcal{L}_{\text{ce}}$ loss is defined as

$$\mathcal{L}_{ce} = -\frac{1}{K_{valid}} \sum_{k \in K_{valid}} \omega_k \cdot \overline{q}_k, \quad (20)$$

where $K_{valid}$ denotes the set of classes present in the current mini batch.

#### 2) Confidence Dispersion Regularization Term

To mitigate prediction instability and under-confidence in minority classes, $\mathcal{L}_{disp}$ penalizes the high dispersion of predicted probabilities within the same class. For class $k$, let $\mu_{p_k}$ and $\sigma_{p_k}$ be the mean and standard deviation of the predicted probabilities $p_{i,k}$ for all samples $i \in \mathcal{B}_k$. Using the Coefficient of Variation (CV) as the regularization term:

$$\mathcal{D}_k = \frac{\sigma_{p_k}}{\mu_{p_k} + \epsilon}. \quad (21)$$

The $\mathcal{L}_{disp}$ loss is the weighted sum of these dispersion terms:

$$\mathcal{L}_{disp} = \sum_{k \in K_{valid}} \frac{N_k}{B} \cdot \omega_k \cdot \mathcal{D}_k. \quad (22)$$

The final training objective is the summation of these two terms:

$$\mathcal{L}_{total} = \mathcal{L}_{ce} + \mathcal{L}_{disp}. \quad (23)$$

By minimizing $\mathcal{L}_{total}$, our model not only adapts to sample scarcity but also encourages consistent and confident predictions within each class, thereby significantly improving performance on tail classes.

### *H. Evaluation Metrics*

To comprehensively evaluate performance, particularly considering the class imbalance inherent in medical datasets, we employed five quantitative metrics: Accuracy, Precision, Macro-average F1-score, Sensitivity, and Specificity. Confusion matrices were further used to analyze class-wise prediction behavior.

## IV. Results

### *A. Dataset*

This study was approved by the Ethics Committee of the Department of Otorhinolaryngology, Second Affiliated Hospital of Xi'an Jiaotong University (No. 2021041). Informed consent was waived due to the retrospective nature of the study and the use of de-identified data

The dataset utilized in this study was retrospectively collected from the Second Affiliated Hospital of Xi'an Jiaotong University. It consists of high-resolution laryngoscopic video sequences from a cohort of 397 patients, including 68 laryngeal mass cases, 94 vocal fold polyp cases, 89 vocal fold nodule cases, 63 vocal fold cyst cases, and 83 normal tissue cases. These videos were acquired using standard clinical stroboscopic laryngoscopes (MATRIX HD3, XION GmbH, Berlin, Germany), capturing multiple cycles of vocal cord vibration. The classification labels for all laryngoscopic videos in our dataset were confirmed by postoperative pathology. Subsequently, these cases were reorganized into three diagnostic categories according to pathological severity. Subjects were grouped according to the clinical type and risk level of the laryngeal lesion. Laryngeal mass cases were categorized as the high-risk group because these lesions require further clinical evaluation and carry a higher concern for neoplastic disease. Vocal fold polyps, vocal fold nodules, and vocal fold cysts, which are benign vocal fold lesions, were categorized as the low-risk group. Subjects without laryngeal lesions were assigned to the normal group. Consequently, the final three-category dataset consisted of 68 high-risk cases, 246 low-risk cases, and 83 normal tissue cases. To ensure assessment robustness and strictly prevent data leakage, we adopted a patient-level [44] stratified partitioning strategy. The 397 participants were randomly assigned to a training set (80%) and a test set (20%). This ensures that all video segments belonging to the same patient appear exclusively in either the training or testing phase.

### *B. Implementation Details*

Our framework was implemented using PyTorch on a computer equipped with a NVIDIA GeForce RTX 2080Ti GPU. For the visual backbone, we utilized the VST-T, initialized with weights pretrained on the Kinetics-400 dataset[45]. The Bi-LSTM in the biomechanical branch consists of two layers with a hidden dimension of 128. To generate the fixed-length input required by the network, we applied a temporal sliding window technique. Each source video was segmented into overlapping clips, with each clip containing T=32 consecutive frames. Consistent with the preprocessing described in Section 3.2, all frames were resized to 224 × 224 pixels and normalized. Correspondingly, 2-dimensional representations of glottal vibration were extracted from the segmentation masks. Ultimately, a total of 1585 video clip samples were generated from the training set, from which 20% were randomly withheld for validation. The network was trained end-to-end using the AdamW optimizer [46] with an initial learning rate of 1e-5 and a cosine annealing learning rate schedule. The batch size was set to 16, and training was conducted for 100 epochs. During the inference phase, to obtain robust video-level predictions from the clip-level model, we employed dense sampling and probabilistic averaging strategy. Specifically, for each variable-length test video, we applied a sliding window with a stride of 32 frames to generate a series of clips covering the entire duration. These clips were individually processed by the network to obtain class probability scores. The final video-level diagnosis was determined by averaging the Softmax probabilities of all constituent clips. This aggregation mechanism ensures that the diagnosis reflects the dominant pathological patterns throughout the examination, mitigating bias from transient artifacts or temporary occlusions.

### *C. Statistical Validation of Biomechanical Priors*

To examine whether the extracted biomechanical priors contain discriminative information for laryngeal lesion classification, we analyzed their distributions across the three diagnostic categories. As illustrated in Fig. 5, we visualized the extracted biomechanical priors in a normalized 9D physical prior space to examine their category-level patterns. In the radar plot, each axis corresponds to one biomechanical descriptor, and each colored polygon represents the normalized category-level profile of the high-risk, low-risk, or normal group. The position of each vertex corresponds to the normalized value of the respective descriptor, with larger radial distances indicating higher values. The radar plot shows that different diagnostic groups exhibit distinct trends in vocal fold dynamics. The high-risk group shows higher gap-related indices and increased ZCR, suggesting more pronounced glottal insufficiency and irregular vibration. In contrast, the normal tissue group generally presents more regular periodicity and stronger bilateral waveform consistency, reflecting relatively stable vocal fold vibration. The low-risk group displays an intermediate pattern, with relatively increased asymmetry but less severe irregularity than the high-risk group. These observations suggest that the extracted biomechanical priors capture meaningful dynamic differences among the three categories and can provide interpretable conditioning information for the subsequent PAGM.

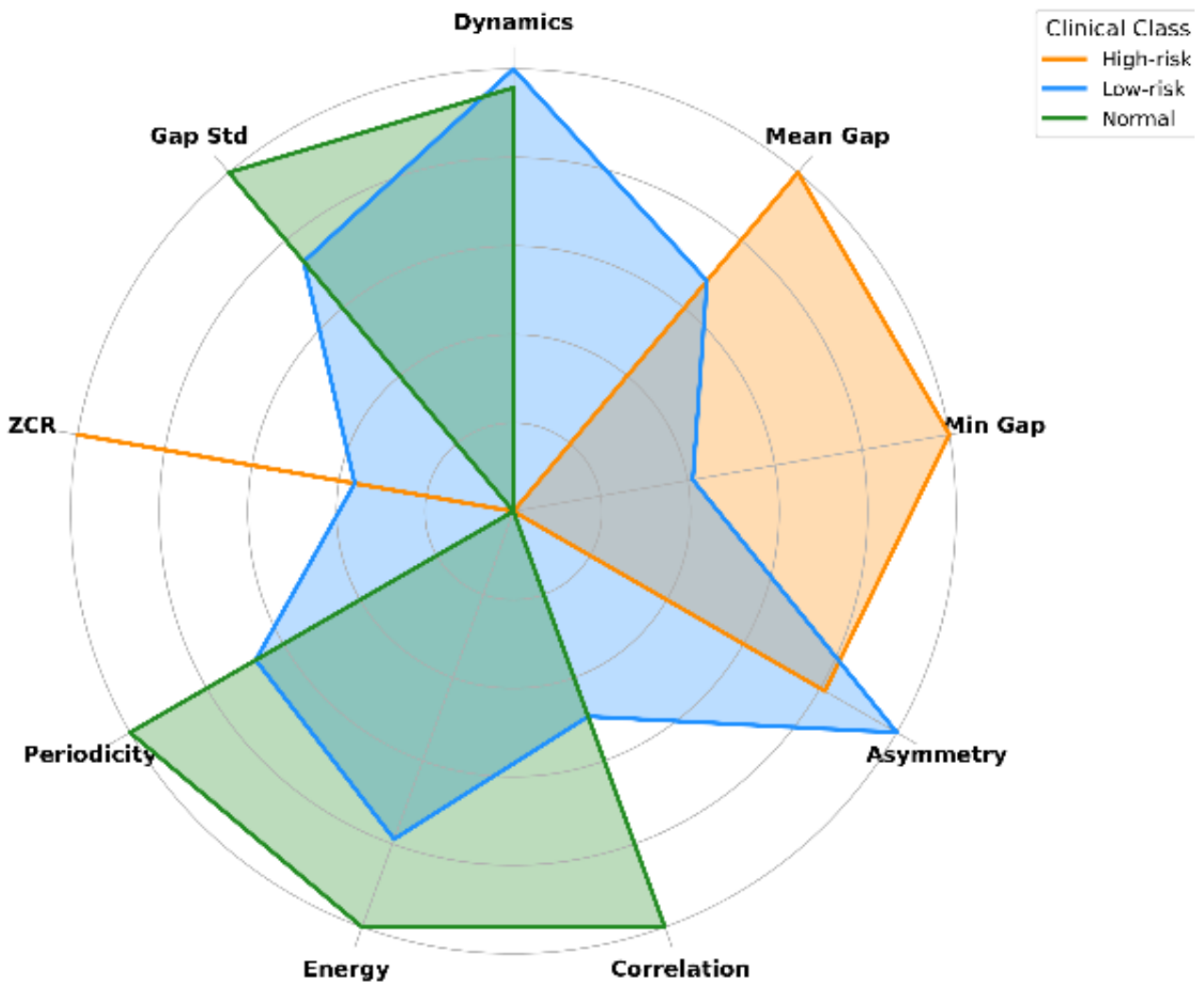

Fig. 5 Clinical Fingerprints of Vocal Fold Pathologies in a 9D Physical Prior Space.

### *D. Comparison Results*

Table I compares PADSNet with representative general-purpose video models from different spatiotemporal modeling paradigms, including X3D [26], Video Swin Transformer[11], VideoMAE [31], V-JEPA2 [32], InternVideo2 [33], SurgMotion [34] and Video Mamba Pro[35]. These baselines cover convolution-based modeling, hierarchical Transformer modeling, self-supervised video representation learning, large-scale video foundation modeling, and state-space modeling, providing a broad evaluation of generic video backbones for laryngoscopic lesion classification.

As shown in Table I, representative video models show limited performance on this clinical task. Although Video Swin Transformer and V-JEPA2 achieve relatively high accuracy of 0.7262, their F1-scores remain limited at 0.5284 and 0.5993, respectively. InternVideo2 obtains the strongest baseline F1-score of 0.6084, while SurgMotion reaches 0.5050, indicating that generic and surgical video foundation representations still struggle to capture vocal fold-specific biomechanical abnormalities. In contrast, PADSNet achieves the best overall performance, with an accuracy of 0.8452 and an F1-score of 0.8089, improving the F1-score by 0.2005 percentage points over the strongest baseline. This demonstrates the advantage of incorporating glottal waveform-derived biomechanical information and physics-aware feature recalibration for laryngeal lesion classification.

Fig. 6 further compares the confusion matrices of different video models on the test set, providing a class-wise analysis of their error patterns. The baseline models exhibit frequent confusion among adjacent categories. This is consistent with the weak visual separability shown in Fig. 1, where diagnostic differences are not always apparent from static appearance alone. Compared with these baselines, PADSNet produces a more balanced confusion pattern across high-risk, low-risk, and normal tissue categories, supporting the effectiveness of incorporating vocal fold biomechanical priors for resolving visually ambiguous lesion boundaries.

TABLE I
Comparison with representative video models

| Model | Accuracy | Precision | Sensitivity | Specificity | F1-score |
|---|---|---|---|---|---|
| X3D[26] | 0.5357 | 0.5233 | 0.5812 | 0.7696 | 0.5218 |
| Video Swin Transformer[11] | 0.7262 | 0.5359 | 0.5491 | 0.7660 | 0.5284 |
| VideoMAE[31] | 0.3929 | 0.5306 | 0.5992 | 0.7431 | 0.4054 |
| InternVideo2[33] | 0.6071 | 0.6038 | 0.6724 | 0.7988 | 0.6084 |
| Video Mamba Pro[35] | 0.4167 | 0.5469 | 0.5936 | 0.7466 | 0.4416 |
| V-JEPA2[32] | 0.7262 | 0.7938 | 0.6046 | 0.7939 | 0.5993 |
| SurgMotion[34] | 0.6548 | 0.6012 | 0.4818 | 0.7364 | 0.5050 |
| Ours | **0.8452** | **0.8451** | **0.7953** | **0.8866** | **0.8089** |

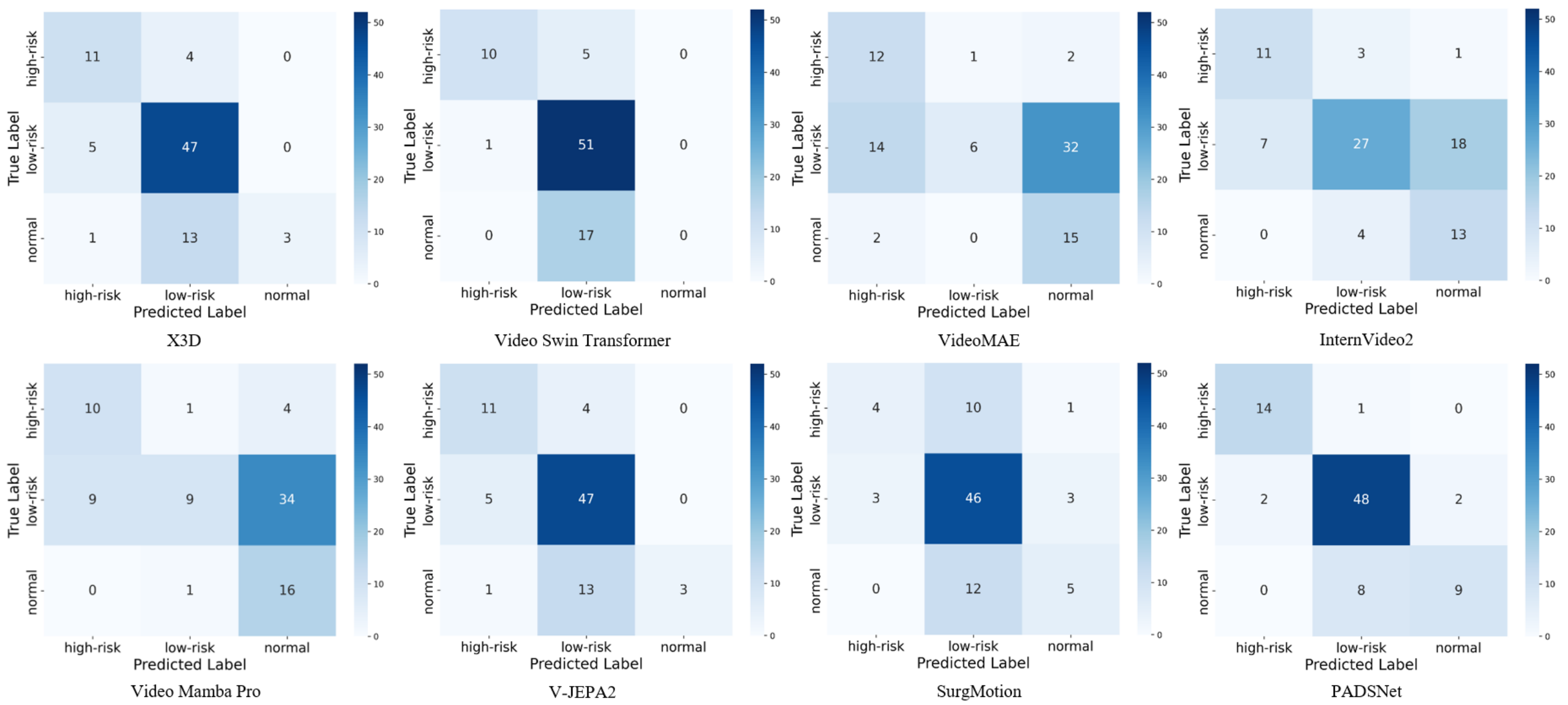

Fig. 6 Comparison of confusion matrices of different video models on the test set.

*E. Ablation Study*

To quantitatively validate the contribution of each component in our proposed framework, we conducted comprehensive ablation experiments. The results are summarized in Table II. We investigated the following configurations: 1) Using only the VST backbone network. 2) Adding the Biomechanical Stream (Bi-LSTM) but fusing features via simple concatenation or averaging. 3) Comparing our WAEM against standard fusion strategies to isolate the benefit of spectral interaction. 4) Employing the proposed WAEM for enhancement but excluding the final PAGM. 5) The complete framework integrating both the WAEM and the PAGM.

As shown in Table II, each component contributes to the final performance. The vision-only VST baseline achieves an accuracy of 0.7262 and an F1-score of 0.5284. Introducing the biomechanical stream improves the accuracy to 0.7738 and the F1-score to 0.6392, indicating that glottal waveform dynamics provide complementary information beyond RGB appearance. Further adding PAGM or WAEM increases the F1-score to 0.7400 and 0.7151, respectively, suggesting that both physics aware recalibration and frequency-aware cross-modal interaction are beneficial. The complete PADSNet achieves the best overall performance, with an accuracy of 0.8452 and an F1-score of 0.8089, demonstrating the complementary effects of WAEM and PAGM in capturing subtle vocal fold vibratory abnormalities.

TABLE II
Ablation experiment results of different modules

| VST | Bi-LSTM | WAEM | PAGM | Accuracy | Precision | Sensitivity | Specificity | F1-score |
|---|---|---|---|---|---|---|---|---|
| √ | | | | 0.7262 | 0.5359 | 0.5491 | 0.7660 | 0.5284 |
| √ | √ | | | 0.7738 | **0.8710** | 0.6486 | 0.8133 | 0.6392 |
| √ | √ | | √ | 0.8095 | 0.8676 | 0.7206 | 0.8501 | 0.7400 |
| √ | √ | √ | | 0.7976 | 0.8634 | 0.7010 | 0.8397 | 0.7151 |
| √ | √ | √ | √ | **0.8452** | 0.8451 | **0.7953** | **0.8866** | **0.8089** |

TABLE III
Effect of different temporal sampling strides

| Stride | Accuracy | Precision | Sensitivity | Specificity | F1-score |
|---|---|---|---|---|---|
| 1 | **0.8452** | **0.8451** | **0.7953** | **0.8866** | **0.8089** |
| 2 | 0.7619 | 0.7212 | 0.7108 | 0.8358 | 0.7026 |
| 3 | 0.7381 | 0.6995 | 0.6954 | 0.8261 | 0.6857 |
| 4 | 0.7262 | 0.6875 | 0.7127 | 0.8372 | 0.7305 |

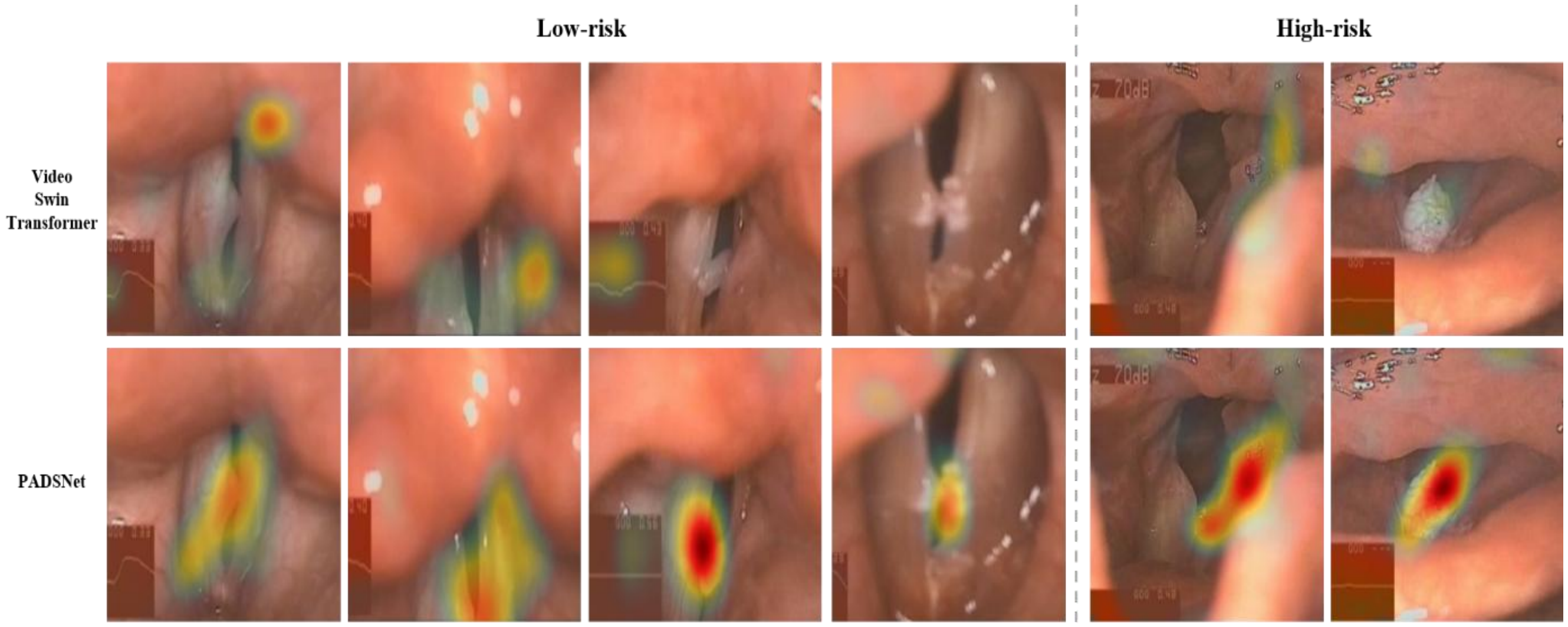

Fig. 7. Comparison of Spatial Attention Maps (CAMs) between Baseline Models and PADSNet.

### *F. Sensitivity Analysis of Sampling Stride*

To evaluate the effect of temporal sampling density, we conducted a sensitivity analysis using different sampling strides. As shown in Table III, stride 1 achieves the best performance, with an accuracy of 0.8452 and an F1-score of 0.8089, indicating that dense sampling is important for preserving vocal fold dynamics. Increasing the stride generally leads to performance degradation, suggesting that sparse sampling may miss transient mucosal wave disturbances, vibratory asymmetry, or subtle glottal closure abnormalities. Although stride 4 obtains a slightly higher F1-score than strides 2 and 3, it remains clearly inferior to stride 1 in overall performance. These results support the use of dense temporal sampling for laryngeal lesion classification.

### *G. Qualitative Analysis and Interpretability*

In medical diagnosis, model interpretability is as paramount as accuracy. To decode the decision-making process of PADSNet, we generated Class Activation Maps (CAMs) to interpret the network's spatial focus (Fig. 7). Unlike baseline models that suffer from artifact-induced attention drift or diffuse glottal focus, PADSNet tends to produce more localized activation around the vocal fold margins and glottal region. This observation provides qualitative evidence that the model attends clinically relevant anatomical regions.

## V. DISCUSSION

This study demonstrates that incorporating biomechanical priors into deep video models is effective for laryngeal lesion classification. The main challenge is the weak inter-class separability in static laryngoscopic views, as diagnostic evidence often resides in subtle vocal fold vibratory abnormalities, such as asymmetry, incomplete closure, or reduced periodicity, rather than in visible morphology. PADSNet addresses this by coupling RGB-based visual features with glottal waveform-derived biomechanical representations, transforming subtle visual differences into measurable dynamic evidence. Ablation results confirm that the biomechanical stream, WAEM, and PAGM each provide complementary contributions.

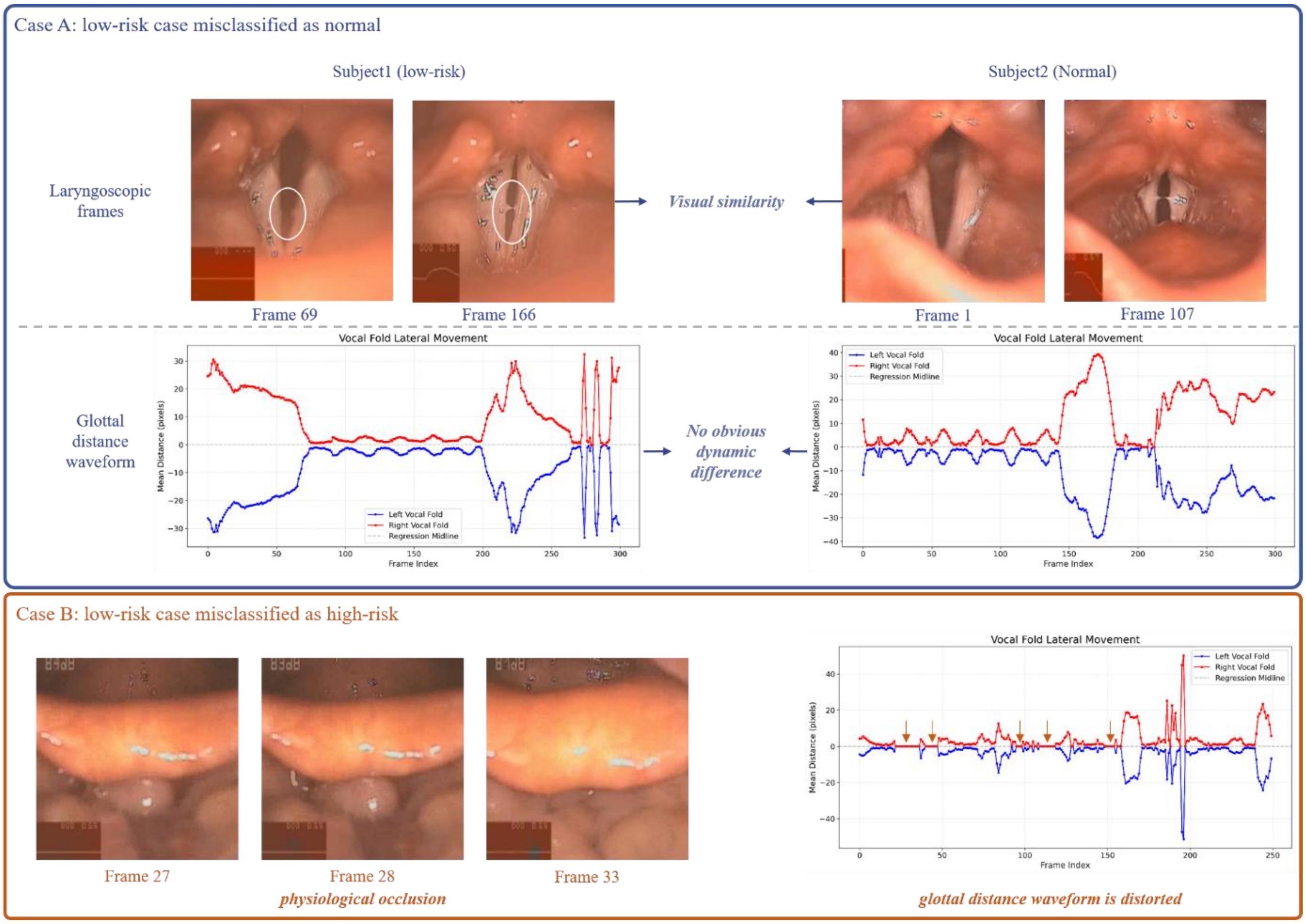

Fig. 8. Case A illustrates a low-risk subject (Subject 1) misclassified as normal. Compared with the normal reference (Subject 2), the low-risk subject exhibits a highly similar laryngoscopic appearance, while the glottal distance waveforms of the two subjects show only limited discriminative differences. Case B demonstrates a low-risk subject misclassified as high-risk. The physiological occlusion disrupts the extraction of the glottal distance waveforms. The affected time intervals correspond to regions where the glottal distance cannot be reliably computed and set to zero, thereby reducing the reliability of the derived biomechanical features. The brown arrows in the waveform plot mark these zero-value intervals caused by the occlusion.

However, several failure cases remain. As illustrated in Fig. 8, most misclassifications occur in borderline cases where adjacent categories exhibit similar static appearances and, in some instances, highly similar glottal waveforms. Case A exemplifies a low-risk subject (Subject 1) misclassified as normal since both the laryngoscopic appearance and the glottal distance waveforms closely resembled those of the normal reference (Subject 2), providing insufficient visual and biomechanical separation. Such cases offer only weak discriminative evidence, making reliable classification difficult even with physics-aware modeling. In addition, some failures are associated with physiological occlusions, which reduce the reliability of the extracted biomechanical priors. Case B illustrates a low-risk subject misclassified as high-risk. During physiological occlusion, the glottal boundaries cannot be reliably segmented, and the corresponding glottal distance values are therefore set to zero. These zero-valued intervals disrupt the waveform, preventing the model from capturing valid vocal-fold dynamics and thereby reducing the reliability of the derived biomechanical features. These cases indicate that the subtle and localized pathological changes still remain challenging.

These failure cases point to a key limitation of the current biomechanical representation. The bivariate glottal-distance waveform and nine scalar priors mainly describe global bilateral motion and may not fully capture spatially localized mucosal-wave propagation. Future work will explore denser and spatially resolved biomechanical representations to better characterize such subtle, localized abnormalities.

Two additional limitations should be noted. First, PADSNet depends on segmentation-derived glottal waveforms, and inaccurate vocal-fold boundaries may propagate errors into both the implicit dynamic features and explicit biomechanical priors. An end-to-end segmentation-classification framework may reduce this dependency. Second, the present study was conducted using a single-center dataset with relatively limited normal-tissue and high-risk samples. External multi-center validation will be required to assess robustness across different devices, acquisition protocols, and patient populations.

## VI. Conclusion

PADSNet provides a physics-aware dual-stream framework for laryngoscopic lesion classification. By integrating RGB-based spatiotemporal appearance features with glottal distance waveform derived biomechanical priors, the proposed method improves the discrimination of high-risk laryngeal masses, low-risk benign lesions, and normal tissue cases. The KCM extracts interpretable dynamic indicators related to vocal fold mobility, symmetry, glottal closure, and vibration stability, while the WAEM and PAGM enable frequency-aware feature interaction and biomechanically guided feature recalibration. Experimental results on a clinical laryngoscopy video dataset demonstrate that PADSNet achieves superior performance over representative video models and provides interpretable evidence for lesion classification. These findings indicate that incorporating explicit vocal fold biomechanical priors can enhance computer-aided laryngoscopic assessment beyond appearance-driven video modeling.

## Acknowledgments

The authors would like to thank the Second Affiliated Hospital of Xi'an Jiaotong University for providing clinical support and laryngoscopic video data for this study. The authors also sincerely appreciate the clinical annotation and diagnostic guidance provided by the otolaryngology team at the same hospital, which were essential for establishing reliable diagnostic labels.

## References

[1] F. Bray, M. Laversanne, H. Sung, et al., "Global cancer statistics 2022: GLOBOCAN estimates of incidence and mortality worldwide for 36 cancers in 185 countries," CA: A Cancer Journal for Clinicians, vol. 74, no. 3, pp. 229–263, May 2024.

[2] B. C. Paul et al., "Diagnostic accuracy of history, laryngoscopy, and stroboscopy," Laryngoscope, vol. 123, no. 1, pp. 215–219, Jan. 2013.

[3] M. P. Karnell et al., "The interrater reliability of stroboscopy evaluations," J. Voice, vol. 26, no. 6, pp. 812.e1–812.e7, Nov. 2012.

[4] M. A. Bruno, E. A. Walker, and H. H. Abujudeh, "Understanding and confronting our mistakes: The epidemiology of error in radiology and strategies for error reduction," RadioGraphics, vol. 35, no. 6, pp. 1668–1676, Oct. 2015, doi: 10.1148/rg.2015150023.

[5] S. Taylor-Phillips and C. Stinton, "Fatigue in radiology: A fertile area for future research," Br. J. Radiol., vol. 92, no. 1099, Art. no. 20190043, Jul. 2019, doi: 10.1259/bjr.20190043.

[6] D. D. Deliyski et al., "Clinical implementation of laryngeal high-speed videoendoscopy: Challenges and evolution," Folia Phoniatr. Logop., vol. 60, no. 1, pp. 33–44, 2008.

[7] R. B. Fujiki, C. K. Croegaert-Koch, and S. L. Thibeault, "Videostroboscopy versus high-speed videoendoscopy: Factors influencing ratings of laryngeal oscillation," J. Speech Lang. Hear. Res., vol. 66, no. 5, pp. 1496–1510, May 2023.

[8] A. Schützenberger, M. Kunduk, M. Döllinger, C. Alexiou, D. Dubrovskiy, M. Semmler, A. Seger, and C. Bohr, "Laryngeal high-speed videoendoscopy: Sensitivity of objective parameters towards recording frame rate," Biomed. Res. Int., vol. 2016, Art. no. 4575437, 2016, doi: 10.1155/2016/4575437.

[9] J. Ren et al., "Automatic recognition of laryngoscopic images using a deep-learning technique," Laryngoscope, vol. 130, no. 11, pp. E686–E693, Nov. 2020.

[10] N. Mohamed et al., "Automated laryngeal cancer detection and classification using dwarf mongoose optimization algorithm with deep learning," Cancers, vol. 16, no. 1, p. 181, Dec. 2023.

[11] Z. Liu et al., "Video swin transformer," in Proc. IEEE/CVF Conf. Comput. Vis. Pattern Recognit. (CVPR), New Orleans, LA, USA, Jun. 2022, pp. 3202–3211.

[12] A. Karpatne et al., "Theory-guided data science: A new paradigm for scientific discovery from data," IEEE Trans. Knowl. Data Eng., vol. 29, no. 10, pp. 2318–2331, Oct. 2017.

[13] N. Esmaeili et al., "Laryngeal lesion classification based on vascular patterns in contact endoscopy and narrow band imaging: Manual versus automatic approach," Sensors, vol. 20, no. 14, p. 4018, Jul. 2020.

[14] C. I. Song et al., "Quantitative evaluation of vocal-fold mucosal irregularities using GLCM-based texture analysis," Laryngoscope, vol. 123, no. 1, pp. E45–E50, Jan. 2013.

[15] G. Andrade-Miranda, Y. Stylianou, D. D. Deliyski, J. I. Godino-Llorente, and N. Henrich Bernardoni, "Laryngeal image processing of vocal folds motion," Appl. Sci., vol. 10, no. 4, p. 1556, Feb. 2020.

[16] N. Dalal and B. Triggs, "Histograms of oriented gradients for human detection," in Proc. IEEE Comput. Soc. Conf. Comput. Vis. Pattern Recognit. (CVPR), San Diego, CA, USA, Jun. 2005, pp. 886–893.

[17] T. Ojala, M. Pietikäinen, and T. Mäenpää, "Multiresolution gray-scale and rotation invariant texture classification with local binary patterns," IEEE Trans. Pattern Anal. Mach. Intell., vol. 24, no. 7, pp. 971–987, Jul. 2002.

[18] D. G. Lowe, "Distinctive image features from scale-invariant keypoints," Int. J. Comput. Vis., vol. 60, no. 2, pp. 91–110, Nov. 2004.

[19] C. Cortes and V. Vapnik, "Support-vector networks," Mach. Learn., vol. 20, no. 3, pp. 273–297, Sep. 1995.

[20] T. Cover and P. Hart, "Nearest neighbor pattern classification," IEEE Trans. Inf. Theory, vol. 13, no. 1, pp. 21–27, Jan. 1967.

[21] D. Voigt, M. Döllinger, T. Braunschweig, A. Yang, U. Eysholdt, and J. Lohscheller, "Classification of functional voice disorders based on phonovibrograms," Artif. Intell. Med., vol. 49, no. 1, pp. 51–59, May 2010, doi: 10.1016/j.artmed.2010.01.001.

[22] W. K. Cho et al., "Diagnostic accuracies of laryngeal diseases using a convolutional neural network-based image classification system," Laryngoscope, vol. 131, no. 11, pp. 2558–2566, Nov. 2021.

[23] B. A. Tran et al., "Support of deep learning to classify vocal fold images in flexible laryngoscopy," Amer. J. Otolaryngol., vol. 44, no. 1, Art. no. 103800, Jan.–Feb. 2023.

[24] R. Wei et al., "A non-local dual-stream fusion network for laryngoscope recognition," *American Journal of Otolaryngology*, vol. 46, 2025, Art. no. 104565, doi: 10.1016/j.amjoto.2024.104565.

[25] M. Qiu et al., "An automatic framework with 3D-large-scale pretrained model for laryngeal cancer detection using laryngoscopic videos," arXiv preprint arXiv:2409.01459, 2024.

[26] C. Feichtenhofer, "X3D: Expanding architectures for efficient video recognition," in Proc. IEEE/CVF Conf. Comput. Vis. Pattern Recognit. (CVPR), Seattle, WA, USA, Jun. 2020, pp. 203–213.

[27] K. Simonyan and A. Zisserman, "Two-stream convolutional networks for action recognition in videos," in Proc. Adv. Neural Inf. Process. Syst. (NIPS), Montreal, QC, Canada, 2014, pp. 568–576.

[28] A. Vaswani, N. Shazeer, N. Parmar, J. Uszkoreit, L. Jones, A. N. Gomez, Ł. Kaiser, and I. Polosukhin, "Attention is all you need," in Proc. Advances in Neural Information Processing Systems (NeurIPS), 2017, pp. 5998–6008.

[29] C. H. J. Kusters et al., "Will Transformers change gastrointestinal endoscopic image analysis? A comparative analysis between CNNs and

Transformers, in terms of performance, robustness and generalization," Medical Image Analysis, 2025, doi: 10.1016/j.media.2024.103348.
[30] K.-N. Wang et al., "TSdetector: Temporal–Spatial self-correction collaborative learning for colonoscopy video detection," Medical Image Analysis, 2025, doi: 10.1016/j.media.2024.103384.
[31] Z. Tong, Y. Song, J. Wang, and L. Wang, "VideoMAE: Masked autoencoders are data-efficient learners for self-supervised video pre-training," in Proc. Adv. Neural Inf. Process. Syst. (NeurIPS), 2022, pp. 10078–10093.
[32] M. Assran, A. Bardes, D. Fan, Q. Garrido, R. Howes, M. Komeili, M. Muckley, A. Rizvi, C. Roberts, K. Sinha, A. Zholus, S. Arnaud, A. Gejji, A. Martin, F. R. Hogan, D. Dugas, P. Bojanowski, V. Khalidov, P. Labatut, F. Massa, M. Szafraniec, K. Krishnakumar, Y. Li, X. Ma, S. Chandar, F. Meier, Y. LeCun, M. Rabbat, and N. Ballas, "V-JEPA 2: Self-supervised video models enable understanding, prediction and planning," arXiv preprint arXiv:2506.09985, 2025.
[33] Y. Wang et al., "InternVideo2: Scaling foundation models for multimodal video understanding," in Proc. Eur. Conf. Comput. Vis. (ECCV), 2024, pp. 3
[34] J. Wu, F. Holm, C. Chen, A. Wang, Y. Hu, X. Ye, Z. Zang, M. Xu, L. Zhou, H. Liao, D. T. M. Chan, M. Feng, W. S. Poon, H. Ren, D. Yi, N. Navab, G. Meng, J. Luo, H. Liu, and Z. Lei, "SurgMotion: A video-native foundation model for universal understanding of surgical videos," arXiv preprint arXiv:2602.05638, 2026.
[35] K. Wang et al., "VideoMambaPro: A leap forward for Mamba in video understanding," arXiv preprint arXiv:2406.18380, 2024.
[36] Y. Bin et al., "Bidirectional long-short term memory for video description," in Proc. ACM Int. Conf. Multimedia (MM), Amsterdam, The Netherlands, Oct. 2016, pp. 436–440.
[37] Pech-Pacheco et al., "Diatom autofocusing in brightfield microscopy: a comparative study," ICPR 2000
[38] F. Isensee, P. F. Jaeger, S. A. A. Kohl, J. Petersen, and K. H. Maier-Hein, "nnU-Net: a self-configuring method for deep learning-based biomedical image segmentation," Nat. Methods, vol. 18, no. 2, pp. 203–211, Feb. 2021.
[39] P. J. Huber, "Robust Estimation of a Location Parameter," Annals of Mathematical Statistics, 1964
[40] L. R. Rabiner and R. W. Schafer, Digital Processing of Speech Signals. Englewood Cliffs, NJ, USA: Prentice-Hall, 1978.
[41] S. Mallat, "A theory for multiresolution signal decomposition: the wavelet representation," IEEE TPAMI, 1989
[42] Vaswani et al., "Attention Is All You Need," NeurIPS 2017
[43] Y. Wang et al., "Adaptive dual-axis style-based recalibration network with class-wise statistics loss for imbalanced medical image classification," IEEE Trans. Med. Imag., vol. 43, no. 3, pp. 984–996, Mar. 2024.
[44] Rouzrokh et al., "Mitigating Bias in Radiology Machine Learning," 2022
[45] W. Kay et al., "The Kinetics human action video dataset," arXiv preprint arXiv:1705.06950, 2017.
[46] Loshchilov and Hutter, "Decoupled Weight Decay Regularization," ICLR 2019